\documentclass[runningheads]{llncs}
\usepackage{graphicx}
\usepackage{amsmath,amssymb} 
\usepackage{dsfont}
\usepackage{color}
\usepackage{lettrine}
\usepackage[width=122mm,left=12mm,paperwidth=146mm,height=193mm,top=12mm,paperheight=217mm]{geometry}
\usepackage{tikz}
\usepackage{multirow}

\usepackage{xspace}
\usepackage{bbm} 
\usepackage{fixltx2e}
\usepackage{contour}
\usepackage{colortbl}

\usepackage[pagebackref=true,breaklinks=true,colorlinks,bookmarks=false]{hyperref} 

\DeclareMathOperator*{\argmin}{\arg\!\min}
\DeclareMathOperator*{\argmax}{\arg\!\max}

\usepackage{tikzextern}
\newcommand{\extfig}[2]{\tikzsetnextfilename{fig/extern/#1}{#2}}
\newcommand{\extdata}[1]{}

\begin{document}
\pagestyle{headings}
\mainmatter
\def\ECCV16SubNumber{825}  

\title{CNN Image Retrieval Learns from BoW:\\ Unsupervised Fine-Tuning with Hard Examples}

\titlerunning{CNN Image Retrieval Learns from BoW} 

\authorrunning{F. Radenovi{\'c}, G. Tolias, and O. Chum} 

\newcommand{\namespace}{\hspace{5mm}} \author{Filip Radenovi{\'c} \namespace Giorgos Tolias \namespace Ond{\v r}ej Chum} 

\institute{CMP, Faculty of Electrical Engineering, Czech Technical University in Prague \\ \email{ \{filip.radenovic,giorgos.tolias,chum\}@cmp.felk.cvut.cz}} 

\maketitle

\def\ie{\emph{i.e.}\xspace}
\def\eg{\emph{e.g.}\xspace}
\def\wrt{\emph{w.r.t.}\xspace}
\def\etal{\emph{et al.}\xspace}

\definecolor{darkred}{rgb}{0.8,0,0}

\newcommand{\real}{\mathbb{R}}
\newcommand{\realnn}{{\mathbb{R}^{+}_{0}}}
\newcommand{\nat}{\mathbb{N}}
\newcommand{\natzero}{{\mathbb{N}_{0}}}

\newcommand{\loss}{\mathcal{L}}

\newcommand{\cX}{\mathcal{X}}
\newcommand{\cI}{\mathcal{I}}
\newcommand{\cP}{\mathcal{P}}
\newcommand{\cE}{\mathcal{E}}
\newcommand{\cN}{\mathcal{N}}
\newcommand{\cM}{\mathcal{M}}

\newcommand{\bG}{\mathbb{G}}
\newcommand{\cG}{\boldsymbol{\mathcal{G}}}

\newcommand{\vf}{\mathbf{f}}
\newcommand{\f}{\mathrm{f}}
\newcommand{\mac}{\bar{\vf}}

\def\l2{$\ell_2$}

\xspaceaddexceptions{+}
\def\cpl2{L\textsubscript{w}\xspace}
\def\pcawhiten{PCA\textsubscript{w}\xspace}

\def\cropI{$\texttt{Crop}_\cI$\xspace}
\def\cropA{$\texttt{Crop}_\cX$\xspace}

\renewcommand{\paragraph}[1]{{\medskip \noindent \bf #1}}
\newcommand{\pari}[1]{{\medskip \noindent \it #1}}
\newcommand{\equ}[1]{Equation~(\ref{#1})\xspace}

\newcommand{\alert}[1]{{\color{red}{#1}}}
\newcommand{\todo}[1]{{\color{blue}{#1}}}

\renewcommand{\b}[1]{\textbf{#1}}
\newcommand{\w}[1]{\color{blue}{#1}}
\newcommand{\ww}[1]{\textbf{\color{blue}{#1}}}

\newcommand{\nb}[1]{\textbf{\color{darkred}{#1}}}
\renewcommand{\sb}[1]{{\color{black}{\contour{darkred}{#1}}}}
\newcommand{\ob}[1]{\textbf{#1}}
\newcommand{\bo}{\cellcolor{gray!15}}

\def\sssp{\hspace{1pt}}
\def\ssp{\hspace{3pt}}
\def\msp{\hspace{5pt}}
\def\bsp{\hspace{8pt}}

\begin{abstract}
Convolutional Neural Networks (CNNs) achieve state-of-the-art performance in many computer vision tasks. However, this achievement is preceded by extreme manual annotation in order to perform either training from scratch or fine-tuning for the target task. In this work, we propose to fine-tune CNN for image retrieval from a large collection of unordered images in a fully automated manner.
We employ state-of-the-art retrieval and Structure-from-Motion (SfM) methods to obtain 3D models, which are used to guide the selection of the training data for CNN fine-tuning. We show that both hard positive and hard negative examples enhance the final performance in particular object retrieval with compact codes.
\end{abstract}

\section{Introduction}
\lettrine{I}{mage} retrieval has received a lot of attention since the advent of invariant local features, such as SIFT~\cite{L04}, and since the seminal work of Sivic and Zisserman~\cite{SZ03} based on Bag-of-Words (BoW). 
Retrieval systems have reached a higher level of maturity by incorporating large visual codebooks~\cite{PCISZ07,AK12}, spatial verification~\cite{PCISZ07,SLBW14} and query expansion~\cite{CMPM11,DGBQG11,TJ14}. 
These ingredients constitute the state of the art on particular object retrieval.
Another line of research focuses on compact image representations in order to decrease memory requirements and increase the search efficiency. 
Representative approaches are Fisher vectors~\cite{PLSP10}, VLAD~\cite{JPDSPS11} and alternatives~\cite{RJC15,AZ13,TFJ14}.
Recent advances~\cite{BSCL14,RSMC14} show that Convolutional Neural Networks (CNN) offer an attractive alternative for image search representations with small memory footprint.

CNNs attracted a lot of attention after the work of Krizhevsky \etal~\cite{KSH12}. Their success is mainly due to the computational power of GPUs and the use of very large annotated datasets~\cite{RDSK+15}. 
Generation of the latter comes at the expense of costly manual annotation. Using CNN layer activations as off-the-shelf image descriptors~\cite{DJVO+13,RASC14} appears very effective and is adopted in many tasks~\cite{GDDM14,IMKG+14,GWGL14}. 
In particular for image retrieval, Babenko \etal~\cite{BSCL14} and Gong \etal~\cite{GWGL14} concurrently propose the use of Fully Connected (FC) layer activations as descriptors, while convolutional layer activations are later shown to have superior performance~\cite{RSMC14,BL15,KMO15,TSJ16}. 

Generalization to other tasks \cite{ARSM+14} is attained by CNN activations, at least up to some extent.
However, initialization by a pre-trained network and re-training for another task, a process called \emph{fine-tuning}, significantly improves the adaptation ability~\cite{ZDGD14,OBLS14}. 
Fine-tuning by training with classes of particular objects, \eg building classes in the work of Babenko \etal~\cite{BSCL14}, is known to improve retrieval accuracy. 
This formulation is much closer to classification than to the desired properties of instance retrieval. 
Typical architectures for metric learning, such as siamese~\cite{CHL05,HCL06,HLT14} or triplet networks~\cite{WSLT+14,SKP15,HA15} employ \emph{matching} and \emph{non-matching} pairs to perform the training and better suit to this task.
In this fashion, Arandjelovic \etal~\cite{AGTPS15} perform fine-tuning based on geo-tagged databases and, similar to our work, they directly optimize the the similarity measure to be used in the final task. 
In contrast to them, we dispense with the need of annotated data or any assumptions on the training dataset. 
A concurrent work~\cite{GARL16} bears resemblance to ours but their focus is on boosting performance through end-to-end learning of a more sophisticated representation, while we target to reveal the importance of hard examples and of training data variation. 

A number of image clustering methods based on local features have been introduced~\cite{CM10a,WL13,PSZ11}. 
Due to the spatial verification, the \emph{clusters} discovered by these methods are reliable. 
In fact, the methods provide not only clusters, but also a matching graph or sub-graph on the cluster images. 
These graphs are further used as an input to a Structure-from-Motion (SfM) pipeline to build a 3D model~\cite{SRCF15}. 
The SfM filters out virtually all mismatched images, and also provides camera positions for all matched images in the cluster. 
The whole process from unordered collection of images to 3D reconstructions is fully automatic.

In this paper, we address an unsupervised fine-tuning of CNN for image retrieval. 
We propose to exploit 3D reconstructions to select the training data for CNN. We show that compared to previous supervised approaches, the variability in the training data from 3D reconstructions delivers superior performance in the image retrieval task. 
During the training process the CNN is trained to learn what a state-of-the-art retrieval system based on local features and spatial verification would match. 
Such a system  has large memory requirements and high query times, while our goal is to mimic this via CNN-based representation.
We derive a short image representation and achieve similar performance to such state-of-the-art systems.

In particular we make the following contributions.
(1) We exploit SfM information and enforce not only hard non-matching (\emph{negative}) but also hard matching (\emph{positive}) examples to be learned by the CNN. This is shown to enhance the derived image representation.
(2) We show that the whitening traditionally performed on short representations~\cite{JC12} is, in some cases, unstable and we rather propose to learn the whitening through the same training data. Its effect is complementary to fine-tuning and it further boosts performance. 
(3) Finally, we set a new state-of-the-art based on compact representations for Oxford Buildings and Paris datasets by re-training well known CNNs, such as AlexNet~\cite{KSH12} and VGG~\cite{SZ14}.
Remarkably, we are on par with existing 256D compact representations even by using 32D image vectors.

\section{Related work}
A variety of previous methods apply CNN activations on the task of image retrieval~\cite{GWGL14,RSMC14,BL15,KMO15,TSJ16,ZZWWT16}.
The achieved accuracy on retrieval is evidence for the generalization properties of CNNs. 
The employed networks were trained for image classification using ImageNet dataset, optimizing classification error.
Babenko \etal~\cite{BSCL14} go one step further and re-train such networks with a dataset that is closer to the target task.
They perform training with object classes that correspond to particular landmarks/buildings. 
Performance is improved on standard retrieval benchmarks.
Despite the achievement, still, the final metric and utilized layers are different to the ones actually optimized during learning.

Constructing such training datasets requires manual effort. 
The same stands for attempts on different tasks~\cite{RASC14,TSJ16} that perform fine-tuning and achieve increase of performance.
In a recent work, geo-tagged datasets with timestamps offer the ground for weakly supervised fine-tuning of a triplet network~\cite{AGTPS15}. 
Two images taken far from each other can be easily considered as non-matching, while matching examples are picked by the most similar nearby images. 
In the latter case, similarity is defined by the current representation of the CNN.
This is the first approach that performs end-to-end fine-tuning for image retrieval and in particular for the task of geo-localization.
The employed training data are now much closer to the final task.
We differentiate by discovering matching and non-matching image pairs in an unsupervised way.
Moreover, we derive matching examples based on 3D reconstruction which allows for harder examples, compared to the ones that the current network identifies. 
Even though hard negative mining is a standard process~\cite{GDDM14,AGTPS15}, this is not the case with hard positive examples. 
Large intra-class variation in classification tasks requires the positive pairs to be sampled carefully;  forcing the model to learn  extremely hard positives may result in over-fitting.
Another exception is the work Simo-Serra \etal~\cite{STFKM14} where they mine hard positive patches for descriptor learning. 
They are also guided by 3D reconstruction but only at patch level.

Despite the fact that one of the recent advances is the triplet loss~\cite{WSLT+14,SKP15,HA15}, note that also Arandjelovic \etal~\cite{AGTPS15} use it, there are no extenstive and direct comparisons to siamese networks and the contrastive loss. 
One exception is the work of Hoffer and Ailon~\cite{HA15}, where triplet loss is shown to be marginally better only on MNIST dataset.
We rather employ a siamese architecture with the contrastive loss and find it to generalize better and to converge at higher performance than the triplet loss.
\section{Network architecture and image representation}
\label{sec:network}
In this section we describe the derived image representation that is based on CNN and we present the network architecture used to perform the end-to-end learning in a siamese fashion.
Finally, we describe how, after fine-tuning, we use the same training data to learn projections that appear to be an effective post-processing step.

\subsection{Image representation}
\begin{figure}[t]
\centering

\def\queryone{11}
\def\dbimageone{3885}
\def\querytwo{42}
\def\dbimagetwo{211}

\setlength{\tabcolsep}{0pt}
\hspace{-20pt}
\begin{tabular}{ccc}
\raisebox{8pt}{\includegraphics[height=65pt]{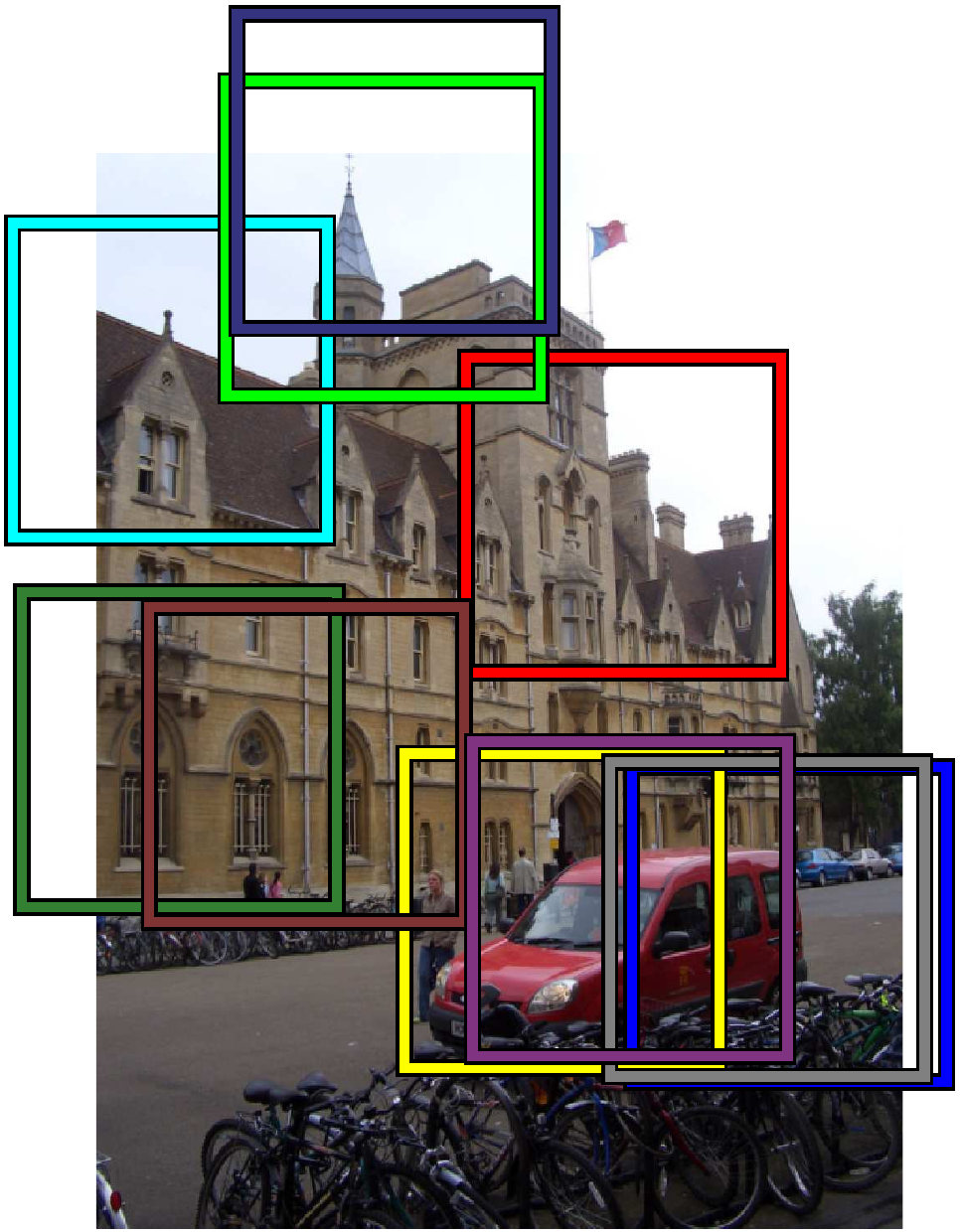} }  &
\includegraphics[height=72pt]{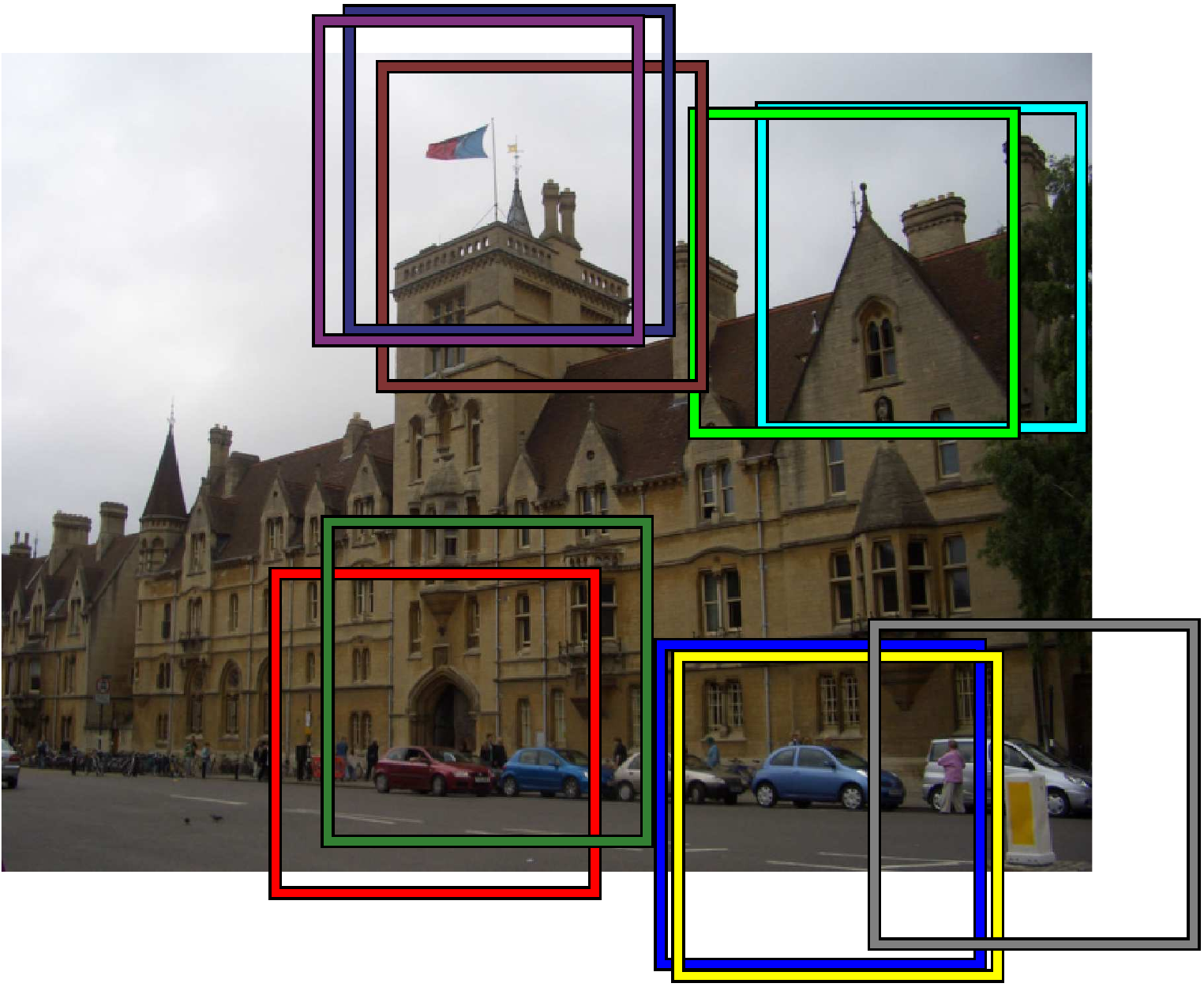}  &
\raisebox{30pt}{
	\begin{tabular}{c}
		\multicolumn{1}{c}{VGG off-the-shelf} \\	
		\foreach \patch in {1,2,3,4,5,6,7,8,9,10}  {
		\includegraphics[height=19pt]{fig/correspondences/q\queryone_db\dbimageone_net0_p\patch_1.png}
		\hspace{-8pt} 
		}\\
		\foreach \patch in {1,2,3,4,5,6,7,8,9,10}  {
		\includegraphics[height=19pt]{fig/correspondences/q\queryone_db\dbimageone_net0_p\patch_2.png} 
		\hspace{-8pt} 
		}\\
	\end{tabular}
}
\\
\includegraphics[height=60pt]{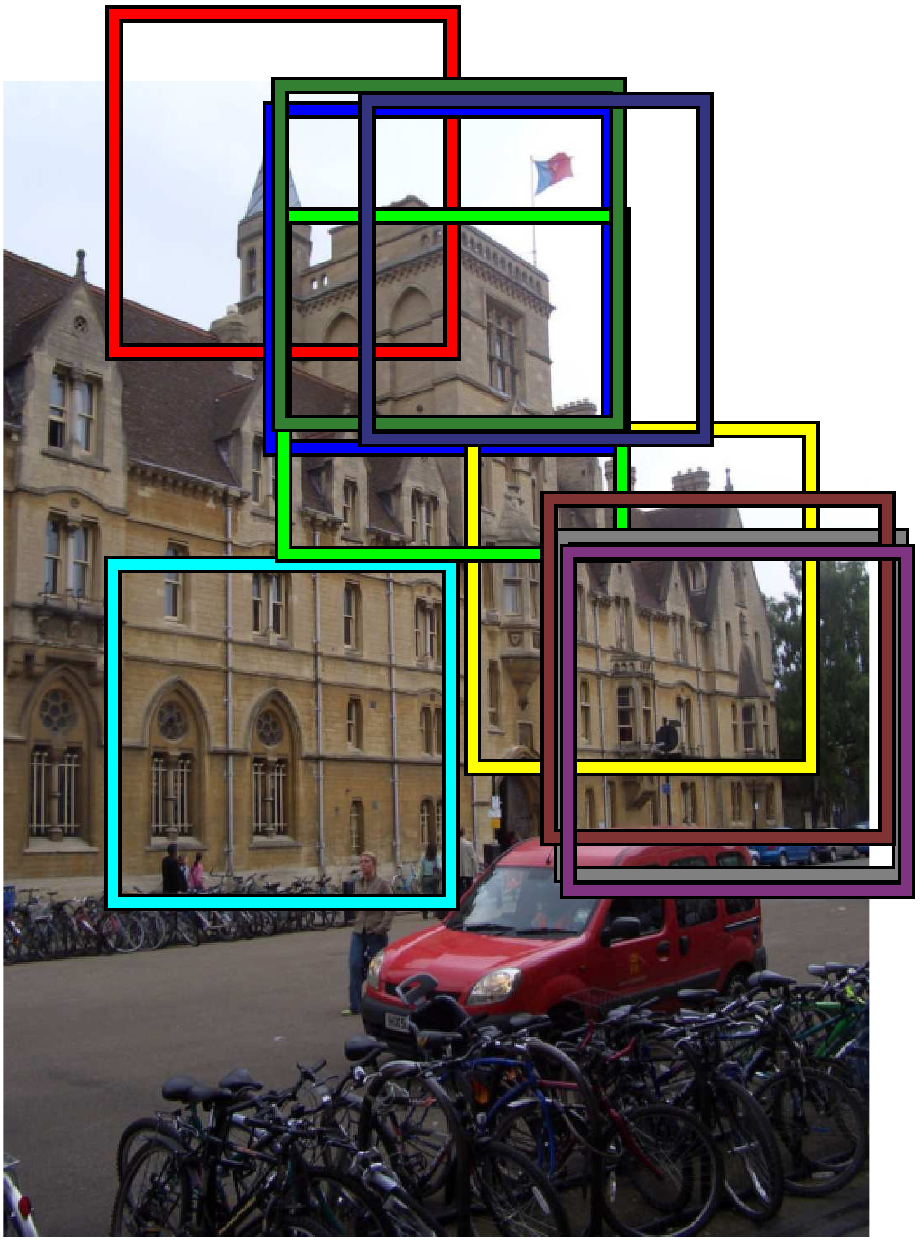} & 
\includegraphics[height=58pt]{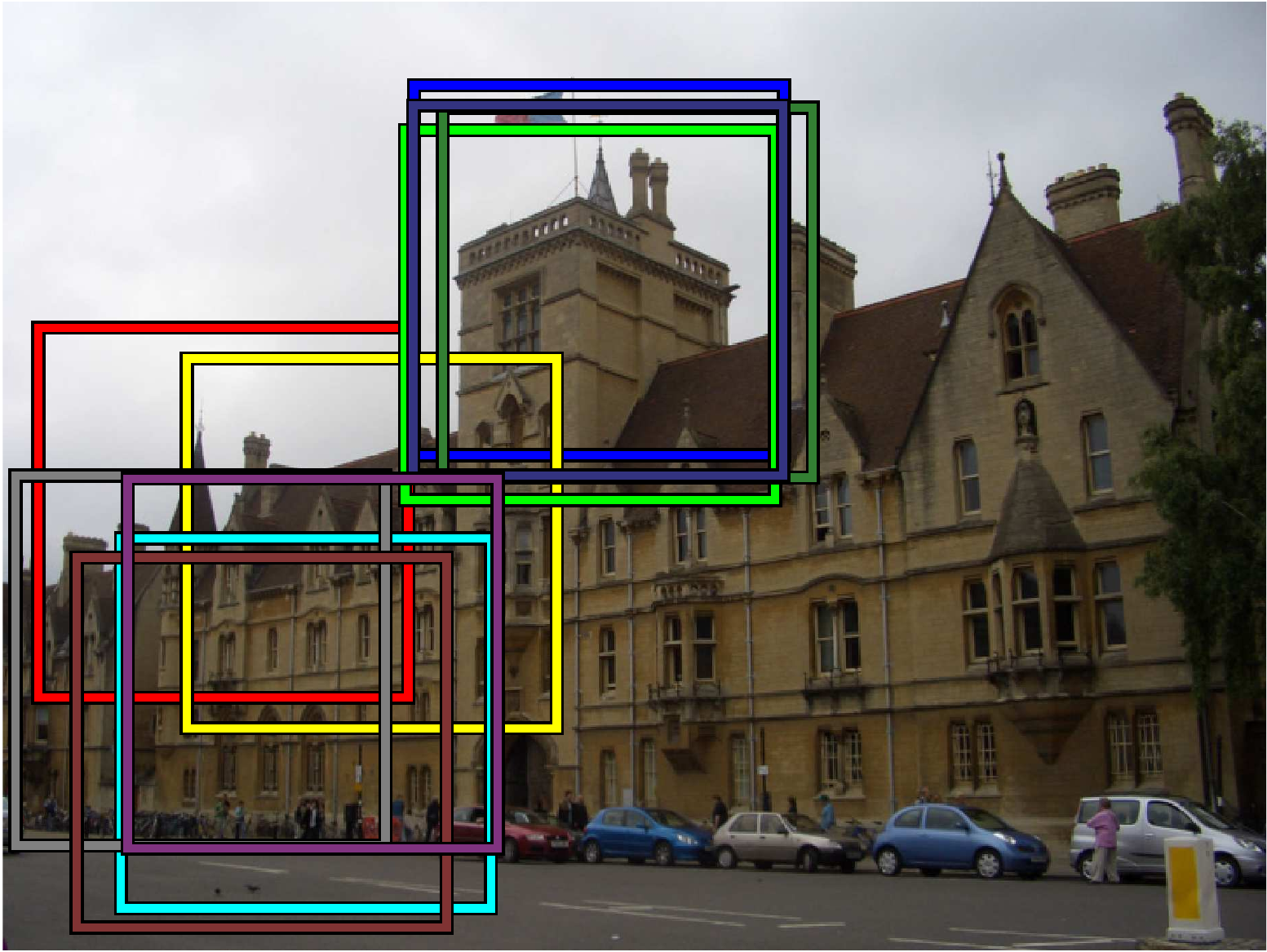} ~~~&
\raisebox{30pt}{
	\begin{tabular}{c}
		\multicolumn{1}{c}{VGG ours} \\	
		\foreach \patch in {1,2,3,4,5,6,7,8,9,10}  {
		\includegraphics[height=19pt]{fig/correspondences/q\queryone_db\dbimageone_net1_p\patch_1.png}
		\hspace{-8pt} 
		}\\
		\foreach \patch in {1,2,3,4,5,6,7,8,9,10}  {
		\includegraphics[height=19pt]{fig/correspondences/q\queryone_db\dbimageone_net1_p\patch_2.png} 
		\hspace{-8pt} 
		}\\
	\end{tabular}
}
\\
\includegraphics[height=70pt]{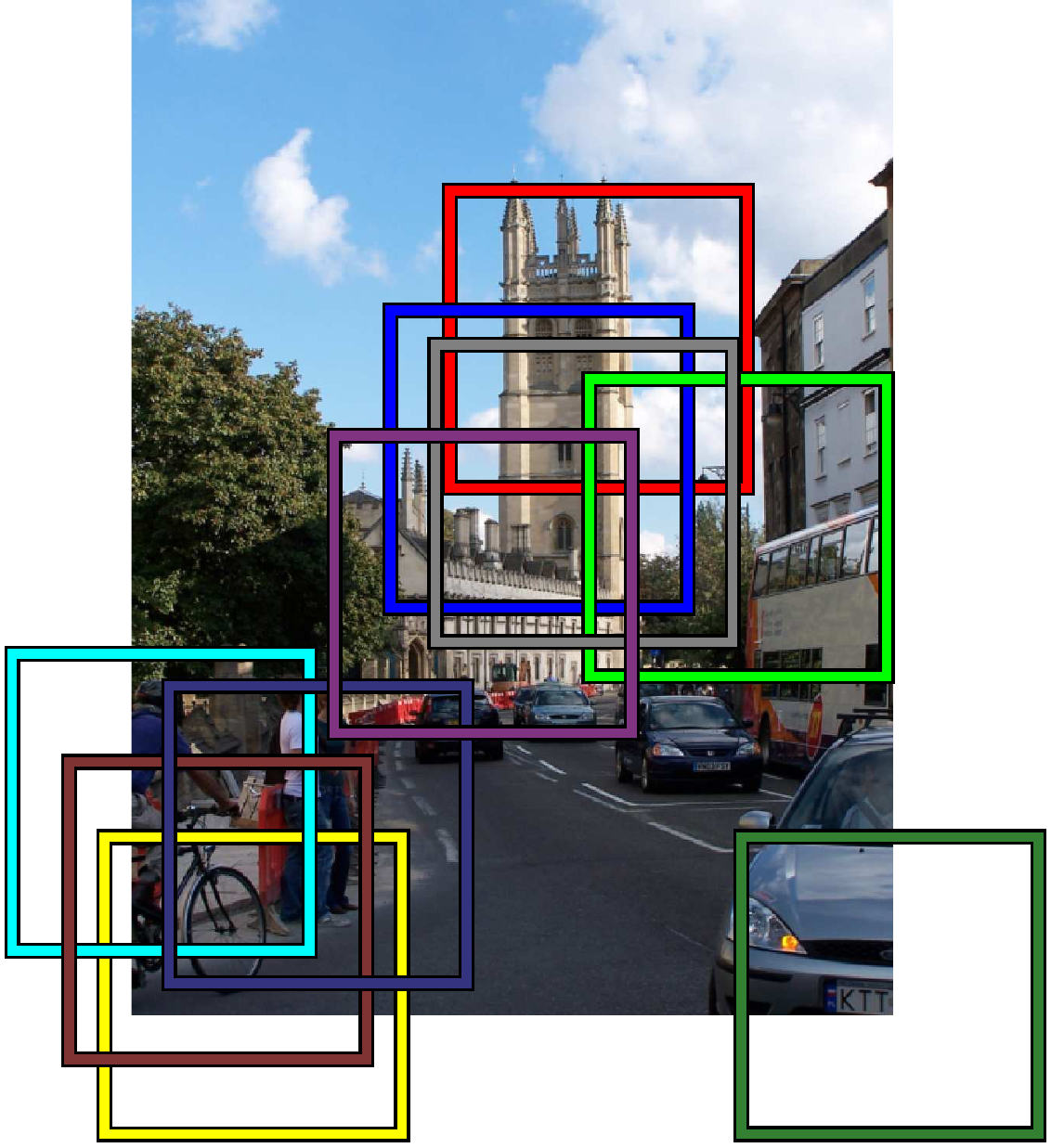} &
\raisebox{8pt}{\includegraphics[height=70pt]{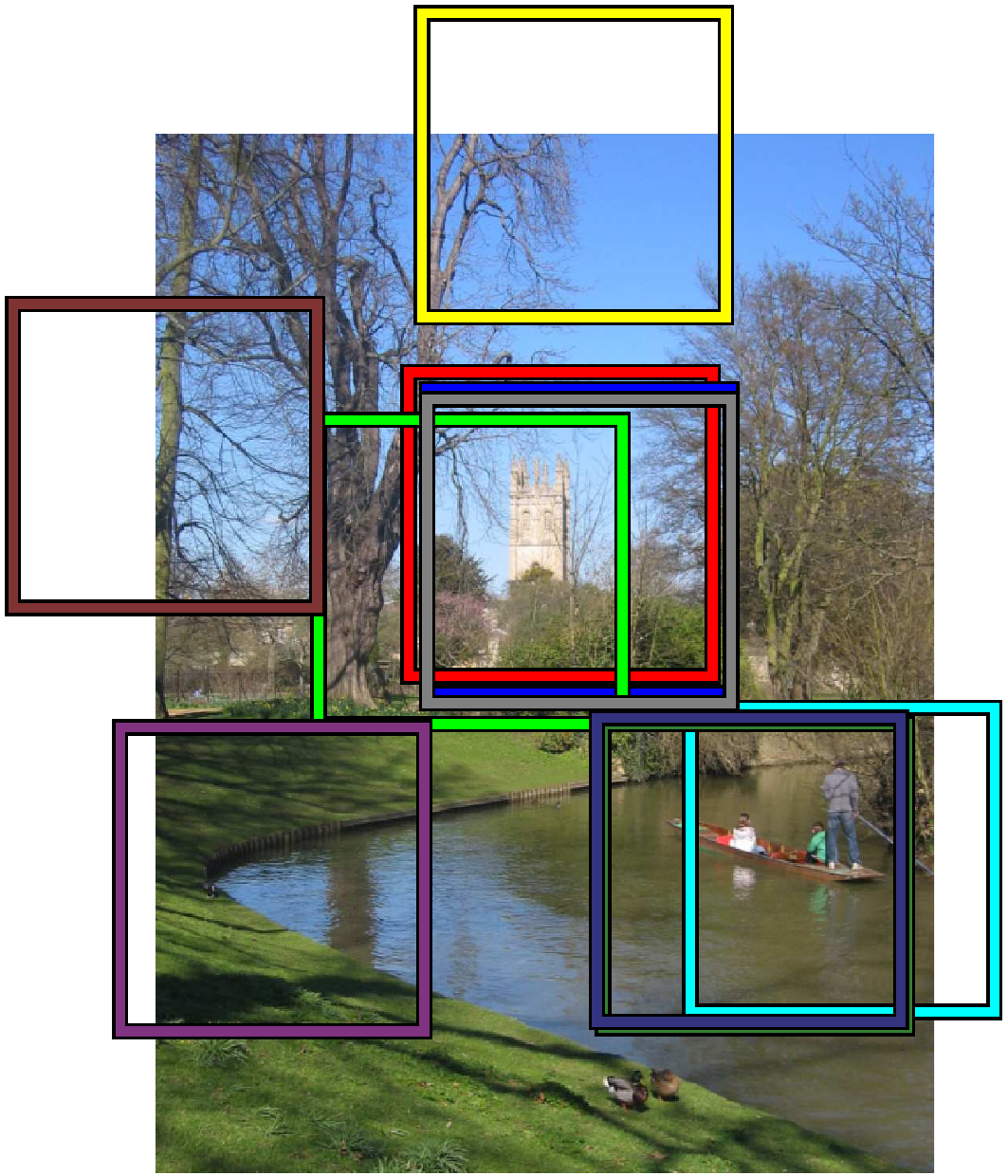}} &
\raisebox{30pt}{
	\begin{tabular}{c}
		\multicolumn{1}{c}{VGG off-the-shelf} \\	
		\foreach \patch in {1,2,3,4,5,6,7,8,9,10}  {
		\includegraphics[height=19pt]{fig/correspondences/q\querytwo_db\dbimagetwo_net0_p\patch_1.png}
		\hspace{-8pt} 
		}\\
		\foreach \patch in {1,2,3,4,5,6,7,8,9,10}  {
		\includegraphics[height=19pt]{fig/correspondences/q\querytwo_db\dbimagetwo_net0_p\patch_2.png} 
		\hspace{-8pt} 
		}\\
	\end{tabular}
}
\\
\includegraphics[height=66pt]{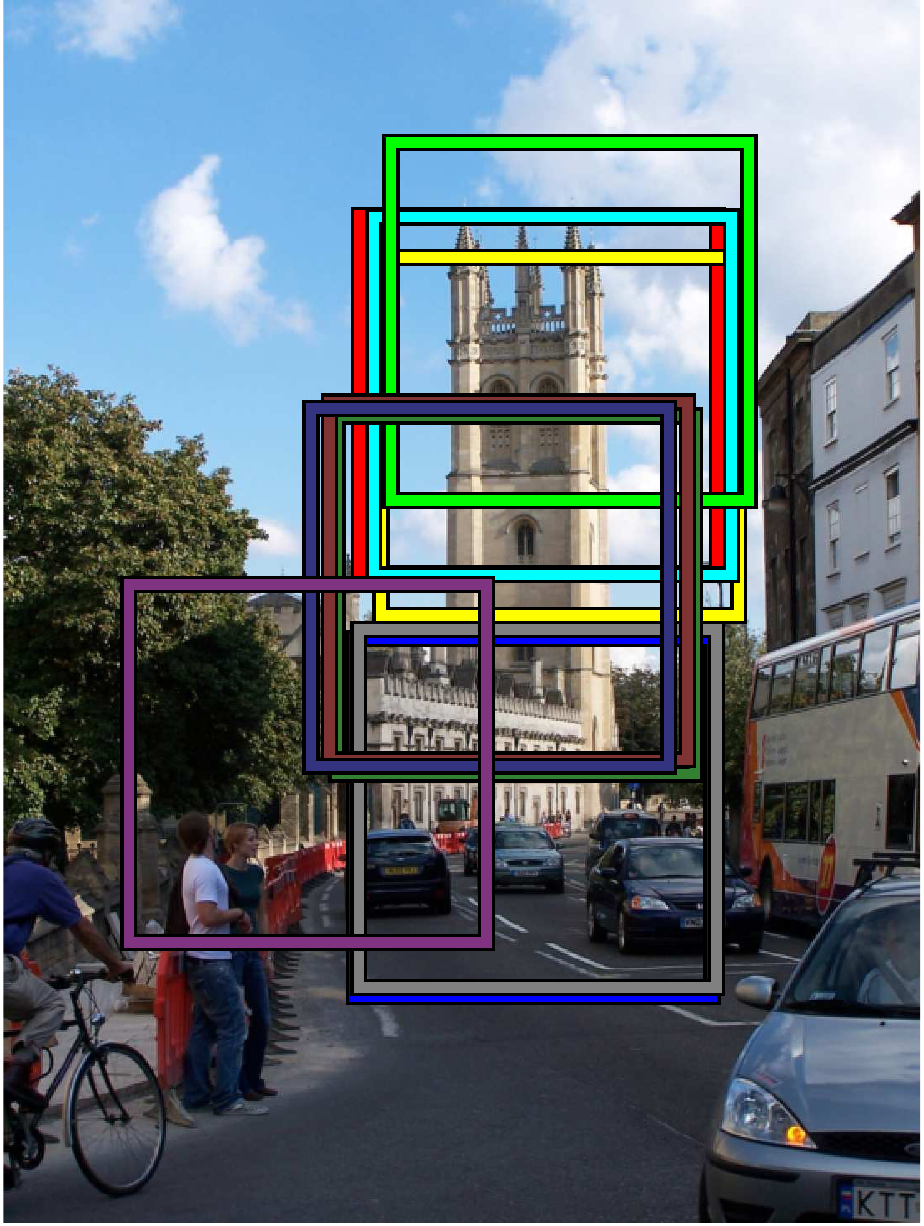} &
~\includegraphics[height=65pt]{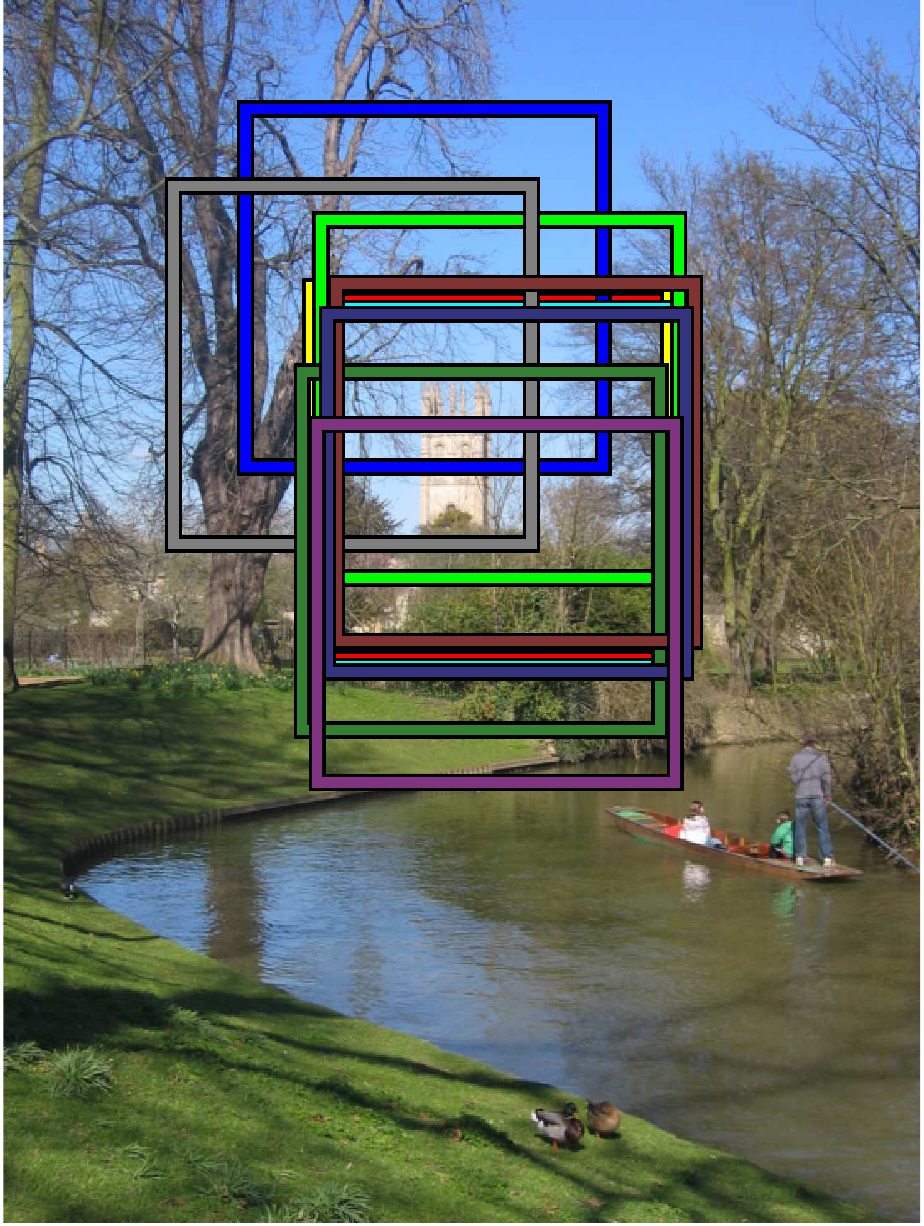} &
\raisebox{30pt}{
	\begin{tabular}{c}
		\multicolumn{1}{c}{VGG ours} \\	
		\foreach \patch in {1,2,3,4,5,6,7,8,9,10}  {
		\includegraphics[height=19pt]{fig/correspondences/q\querytwo_db\dbimagetwo_net1_p\patch_1.png}
		\hspace{-8pt} 
		}\\
		\foreach \patch in {1,2,3,4,5,6,7,8,9,10}  {
		\includegraphics[height=19pt]{fig/correspondences/q\querytwo_db\dbimagetwo_net1_p\patch_2.png} 
		\hspace{-8pt} 
		}\\
	\end{tabular}
}
\\
\end{tabular}
\vspace{-10pt}
\caption{Visualization of patches corresponding to the MAC vector components that have the highest contribution to the pairwise image similarity. Examples shown use CNN before (top) and after (bottom) fine-tuning of VGG. The same color corresponds to the same vector component (feature map) per image pair. The patch size is equal to the receptive field of the last pooling layer.
\label{fig:mac_matches}
\vspace{-15pt}}
\end{figure}

\vspace{-2pt}
We adopt a compact representation that is derived from activations of convolutional layers and is shown to be effective for particular object retrieval~\cite{ARSM+14,TSJ16}.
We assume that a network is fully convolutional~\cite{PKS15} or that all fully connected layers are discarded.
Now, given an input image, the output is a 3D tensor $\cX$ of $W\times H \times K$ dimensions, where $K$ is the number of feature maps in the last layer. 
Let $\cX_k$ be the set of all $W\times H$ activations for feature map $k \in \{1 \ldots K$\}.
The network output consists of $K$ such sets of activations.
The image representation, called Maximum Activations of Convolutions (MAC)~\cite{RSMC14,TSJ16}, is simply constructed by max-pooling over all dimensions per feature map and is given by
\vspace{-2pt}
\begin{equation}
\vspace{-2pt}
\vf = [\f_1 \ldots \f_k \ldots \f_K]^\top \text{,~with~} \f_k = \max_{x\in \cX_{k}}~x \cdot \mathds{1}(x>0).
\label{equ:mac}
\end{equation}
\noindent
The indicator function $\mathds{1}$ takes care that the feature vector $\vf$ is non-negative, as if the last network layer was a Rectified Linear Unit (ReLU).
The feature vector finally consists of the maximum activation per feature map and its dimensionality  is equal to $K$.
For many popular networks this is equal to 256 or 512, which makes it a compact image representation.
MAC vectors are subsequently \l2-normalized and similarity between two images is evaluated with inner product. 
The contribution of a feature map to the image similarity is measured by the product of the corresponding MAC vector components. 
In Figure~\ref{fig:mac_matches} we show the image patches in correspondence that contribute most to the similarity. 
Such implicit correspondences are improved after fine-tuning. Moreover, the CNN fires less to ImageNet classes, \eg cars and bicycles. 
%
%
\vspace{-6pt}
\subsection{Network and siamese learning}
The proposed approach is applicable to any CNN that consists of only convolutional layers. 
In this paper, we focus on re-training (\ie fine-tuning) state-of-the-art CNNs for classification, in particular AlexNet and VGG. 
Fully connected layers are discarded and the pre-trained networks constitute the initialization for our convolutional layers.
Now, the last convolutional layer is followed by a MAC layer that performs MAC vector computation (\ref{equ:mac}).
The input of a MAC layer is a 3D tensor of activation and the output is a non-negative vector. 
Then, an \l2-normalization block takes care that output vectors are normalized. 
In the rest of the paper, MAC corresponds to the \l2-normalized vector $\mac$.

We adopt a siamese architecture and train a two branch network. 
Each branch is a clone of the other, meaning that they share the same parameters. 
Training input consists of image pairs $(i,j)$ and labels $Y(i,j)\in \{0, 1\}$ declaring whether a pair is non-matching (label 0) or matching (label 1). 
We employ the contrastive loss~\cite{CHL05} that acts on the (non-)matching pairs and is defined as
\begin{equation}
\small
\loss(i,j) = \frac{1}{2}\left(Y(i,j) ||\mac(i)-\mac(j)||^2 + \left(1-Y(i,j)\right) \left(\max\{0, \tau - ||\mac(i)-\mac(j)||\}\right)^2\right),
\end{equation}
where $\mac(i)$ is the \l2-normalized MAC vector of image $i$, and $\tau$ is a parameter defining when non-matching pairs have large enough distance in order not to be taken into account in the loss.
We train the network using Stochastic Gradient Descent (SGD) and a large training set created automatically (see Section~\ref{sec:dataset}). 
\subsection{Whitening and dimensionality reduction}
\label{ref:projections}
\vspace{-5pt}
In this section, the post-processing of fine-tuned MAC vectors is considered. 
Previous methods~\cite{BL15,TSJ16} use PCA of an independent set for whitening and dimensionality reduction, that is the covariance matrix of all descriptors is analyzed. We propose to take advantage of the labeled data provided by the 3D models and use linear discriminant projections originally proposed by Mikolajczyk and Matas~\cite{MM07}. The projection is decomposed into two parts, whitening and rotation. 
The whitening part is the inverse of the square-root of the intraclass (matching pairs) covariance matrix $C_S^{-\frac{1}{2}}$, where 
\vspace{-6pt}
\begin{equation}
C_S = \sum_{Y(i,j)=1} \left(\mac(i) - \mac(j)\right)\left(\mac(i) - \mac(j)\right)^\top.
\end{equation}
The rotation part is the PCA of the interclass (non-matching pairs) covariance matrix in the whitened space $\mathrm{eig}(C_S^{-\frac{1}{2}} C_D C_S^{-\frac{1}{2}})$, where 
\vspace{-6pt}
\begin{equation}
C_D = \sum_{Y(i,j)=0} \left(\mac(i) - \mac(j)\right)\left(\mac(i) - \mac(j)\right)^\top.
\end{equation}
The projection $P = C_S^{-\frac{1}{2}} \mathrm{eig}(C_S^{-\frac{1}{2}} C_D C_S^{-\frac{1}{2}})$ is then applied as $P^\top (\mac(i)-\mu)$, where $\mu$ is the mean MAC vector to perform centering. To reduce the descriptor dimensionality to $D$ dimensions, only eigenvectors corresponding to $D$ largest eigenvalues are used.
Projected vectors are subsequently \l2-normalized.
\section{Training dataset}
\label{sec:dataset}
In this section we briefly summarize the tightly-coupled BoW and SfM reconstruction system~\cite{SRCF15,RSJFCM16} that is employed to automatically select our training data. 
Then, we describe how we exploit the 3D information to select harder matching pairs and hard non-matching pairs with larger variability. 

\subsection{BoW and 3D reconstruction}
The retrieval engine used in the work of Schonberger \etal~\cite{SRCF15} builds upon BoW with fast spatial verification~\cite{PCISZ07}. 
It uses Hessian affine local features~\cite{MTSZMSKG05}, \mbox{RootSIFT} descriptors~\cite{AZ12}, and a fine vocabulary of 16M visual words~\cite{MPCM13}.
Then, query images are chosen via min-hash and spatial verification, as in~\cite{CM10a}. 
Image retrieval based on BoW is used to collect images of the objects/landmarks.
These images serve as the initial matching graph for the succeeding SfM reconstruction, which is performed using state-of-the-art SfM~\cite{FGGJR10,AFSS+11}. Different mining techniques, \eg zoom in, zoom out~\cite{MCM13,MRCM14}, sideways crawl~\cite{SRCF15}, help to build larger and complete model. 

In this work, we exploit the outcome of such a system. 
Given a large unannotated image collection, images are clustered and a 3D model is constructed per cluster.
We use the terms \emph{3D model}, \emph{model} and \emph{cluster} interchangeably.
For each image, the estimated camera position is known, as well as the local features registered on the 3D model. 
We drop redundant (overlapping) 3D models, that might have been constructed from different seeds.
Models reconstructing the same landmark but from different and disjoint viewpoints are considered as non-overlapping.

\subsection{Selection of training image pairs}

A 3D model is described as a bipartite visibility graph $\bG = (\cI \cup \cP,\cE)$~\cite{LSH10}, where images $\cI$ and points $\cP$ are the vertices of the graph. 
Edges of this graph are defined by visibility relations between cameras and points, \ie if a point $p\in \cP$ is visible in an image $i\in \cI$, then there exists an edge $(i,p) \in \cE$. 
The set of points observed by an image $i$ is given by
\begin{equation}
\label{equ:observed_points}
\cP(i) = \{ p \in \cP: (i,p) \in \cE \}.
\end{equation} 

We create a dataset of tuples $\left(q, m(q), \cN(q)\right)$, where $q$ represents a query image, $m(q)$ is a positive image that matches the query, and $\cN(q)$ is a set of negative images that do not match the query.
These tuples are used to form training image pairs, where each tuple corresponds to $|\cN(q)|+1$ pairs. 
For a query image $q$, a pool $\cM(q)$ of candidate positive images is constructed based on the camera positions in the cluster of $q$.
It consists of the $k$ images with closest camera centers to the query.
Due to the wide range of camera orientations, these do not necessarily depict the same object. 
We therefore propose three different ways to sample the positive image.
The positives examples are fixed during the whole training process for all three strategies.

\paragraph{Positive images: MAC distance.} 
The image that has the lowest MAC distance to the query is chosen as positive, formally
\begin{equation}
m_1(q) = \argmin_{i \in \cM(q)} ||\mac(q)-\mac(i)||.
\label{equ:mac_pos}
\end{equation} 
This strategy is similar to the one followed by Arandjelovic \etal~\cite{AGTPS15}. 
They adopt this choice since only GPS coordinates are available and not camera orientations.
Downside of this approach is that the chosen matching examples already have low distance, thus not forcing network to learn much out of the positive samples.

\paragraph{Positive images: maximum inliers.} 
In this approach, the 3D information is exploited to choose the positive image, independently of the CNN descriptor. In particular, the image that has the highest number of co-observed 3D points with the query is chosen.
That is, 
\begin{equation}
\label{equ:ninl_pos}
m_2(q) = \argmax_{i \in \cM(q)} |\cP(q) \cap \cP(i)|.
\end{equation} 
This measure corresponds to the number of spatially verified features between two images, a measure commonly used for ranking in BoW-based retrieval. As this choice is independent of the CNN representation, it delivers more challenging positive examples.
\begin{figure}[t]
\centering

\def\imheight{1.1}
\def\qnumone{1021}
\def\qnumtwo{1441}
\def\qnumthree{2461}
\def\qnumfour{6511}
\def\qnumfive{3481}
\def\qnumsix{3991}
\def\qnumseven{1111}
\def\qnumeight{5311}

\def\vnum{5}
\def\raisenum{0}

\setlength{\fboxsep}{0pt}%
\setlength{\fboxrule}{2pt}%

\setlength\tabcolsep{1.5mm}

\begin{tabular}{lclc}
\fcolorbox{green}{black}{\includegraphics[height=\imheight cm]{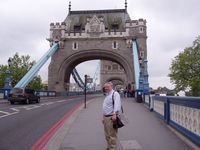}} &
\includegraphics[height=\imheight cm]{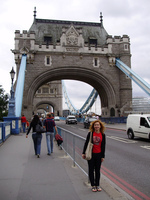} 
\includegraphics[height=\imheight cm]{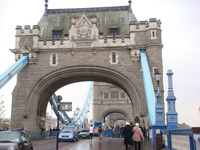} 
\includegraphics[height=\imheight cm]{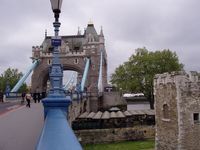}
&
\fcolorbox{green}{black}{\includegraphics[height=\imheight cm]{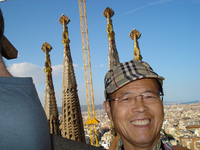}} &
\includegraphics[height=\imheight cm]{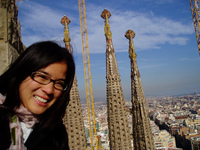} 
\includegraphics[height=\imheight cm]{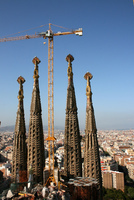} 
\includegraphics[height=\imheight cm]{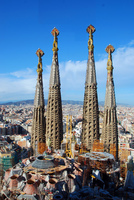}
\\
\fcolorbox{green}{black}{\includegraphics[height=\imheight cm]{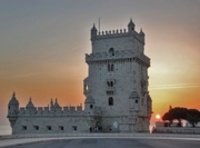}} & 
\includegraphics[height=\imheight cm]{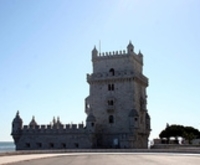} 
\includegraphics[height=\imheight cm]{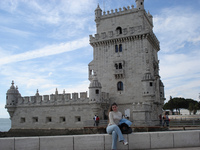} 
\includegraphics[height=\imheight cm]{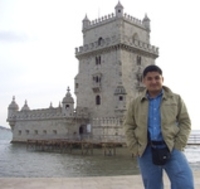}
&
\fcolorbox{green}{black}{\includegraphics[height=\imheight cm]{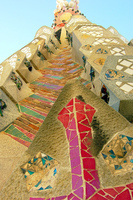}} & 
\includegraphics[height=\imheight cm]{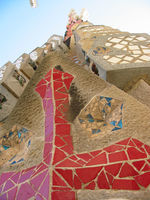} 
\includegraphics[height=\imheight cm]{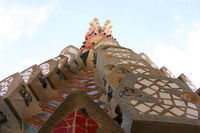} 
\includegraphics[height=\imheight cm]{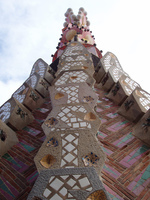}
\\
\fcolorbox{green}{black}{\includegraphics[height=\imheight cm]{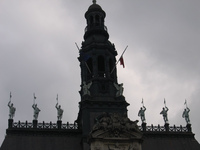}} & 
\includegraphics[height=\imheight cm]{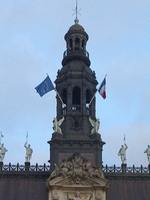} 
\includegraphics[height=\imheight cm]{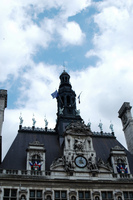} 
\includegraphics[height=\imheight cm]{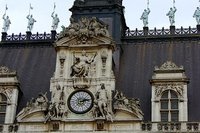}
&
\fcolorbox{green}{black}{\includegraphics[height=\imheight cm]{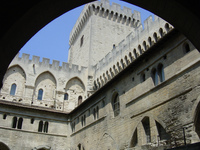}} & 
\includegraphics[height=\imheight cm]{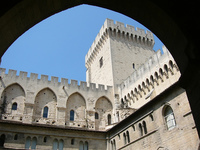} 
\includegraphics[height=\imheight cm]{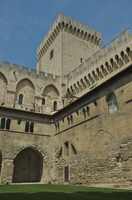} 
\includegraphics[height=\imheight cm]{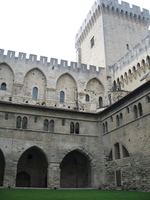}
\\
\fcolorbox{green}{black}{\includegraphics[height=\imheight cm]{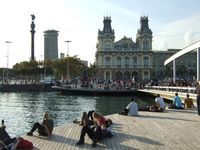}} & 
\includegraphics[height=\imheight cm]{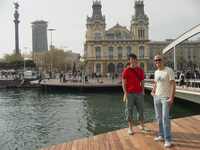} 
\includegraphics[height=\imheight cm]{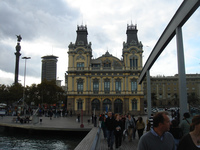} 
\includegraphics[height=\imheight cm]{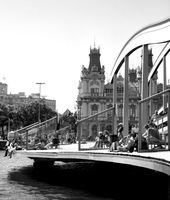}
&
\fcolorbox{green}{black}{\includegraphics[height=\imheight cm]{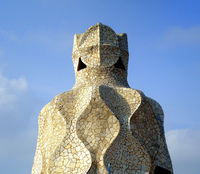}} & 
\includegraphics[height=\imheight cm]{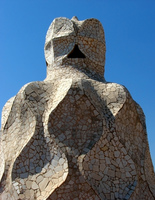} 
\includegraphics[height=\imheight cm]{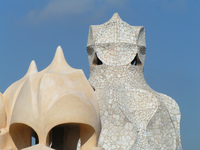} 
\includegraphics[height=\imheight cm]{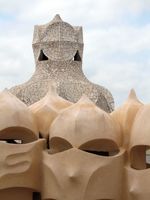}
\\

\end{tabular}
\caption{Examples of training query images (green border) and matching images selected as positive examples by methods (from left to right) $m_1(q)$, $m_2(q)$, and $m_3(q)$.
\label{fig:positives}}
\end{figure}

\paragraph{Positive images: relaxed inliers.}
Even though both previous methods choose positive images depicting the same object as the query, the variance of viewpoints is limited.
Instead of using a pool of images with similar camera position, the positive example is selected at random from a set of images that co-observe enough points with the query, but do not exhibit too extreme scale change. 
The positive example in this case is  
\begin{equation}
\label{equ:relaxed_pos}
m_3(q) = \texttt{random}\left\{ i \in \cM(q): \frac{|\cP(i) \cap \cP(q)|}{|\cP(q)|} \geq t_i,~\texttt{scale}(i,q) \leq t_s \right\},
\end{equation} 
where $\texttt{scale}(i,q)$ is the scale change between the two images.
This method results in selecting harder matching examples which are still guaranteed to depict the same object. Method $m_3$ chooses different image than $m_1$ on 86.5\% of the queries.
In Figure~\ref{fig:positives} we present examples of query images and the corresponding positives selected with the three different methods. The relaxed method increases the variability of viewpoints. 

\begin{figure}[t]
\centering

\def\imheight{1cm}
\def\qnumone{253}
\def\qnumtwo{1611}
\def\qnumthree{1609}

\hspace{-10pt}

\setlength\tabcolsep{2mm}

\begin{tabular}{cccc}


\fcolorbox{green}{black}{\includegraphics[height=\imheight]{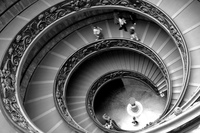}} &
\includegraphics[height=\imheight]{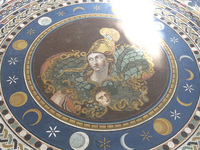} &
\foreach \neg in {2,3}  { 
	\includegraphics[height=\imheight]{fig/negatives/q\qnumone_m2_n\neg.jpg}
} $\ldots$& 
\foreach \neg in {4,5}  { 
	\includegraphics[height=\imheight]{fig/negatives/q\qnumone_m1_n\neg.jpg}
}$\ldots$\\


\fcolorbox{green}{black}{\includegraphics[height=\imheight]{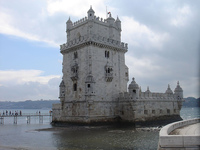}} &
\includegraphics[height=\imheight]{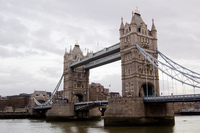} &
\foreach \neg in {3,4}  { 
	\includegraphics[height=\imheight]{fig/negatives/q\qnumtwo_m2_n\neg.jpg}
}$\ldots$ & 
\foreach \neg in {2,5}  { 
	\includegraphics[height=\imheight]{fig/negatives/q\qnumtwo_m1_n\neg.jpg}
}$\ldots$\\


\fcolorbox{green}{black}{\includegraphics[height=\imheight]{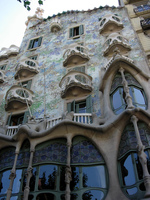}} &
\includegraphics[height=\imheight]{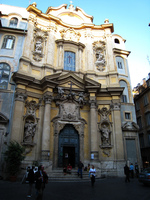} &
\foreach \neg in {2,3,4}  { 
	\includegraphics[height=\imheight]{fig/negatives/q\qnumthree_m2_n\neg.jpg}
}$\ldots$ & 
\foreach \neg in {2,3,4}  { 
	\includegraphics[height=\imheight]{fig/negatives/q\qnumthree_m1_n\neg.jpg}
}$\ldots$\\

$q$ & $n(q)$ & $\cN_1(q)$ & $\cN_2(q)$ \\


\end{tabular}\\[-.7\baselineskip]
\vspace{5pt}
\caption{Examples of training query images $q$ (green border), hardest non-matching images $n(q)$ that are always selected as negative examples, and additional non-matching images selected as negative examples by $\cN_1(q)$ and $\cN_2(q)$ methods respectively.
\label{fig:negatives}
\vspace{10pt}}
\end{figure}

\paragraph{Negative images.} 
Negative examples are selected from clusters different than the cluster of the query image, as the clusters are non-overlaping. 
Following a well-known procedure, we choose hard negatives~\cite{STFKM14,GDDM14}, that is, non-matching images with the most similar descriptor. Two different strategies are proposed. In the first, $\cN_1(q)$, k-nearest neighbors from all non-matching images are selected. In the other, $\cN_2(q)$, the same criterion is used, but at most one image per cluster is allowed. While $\cN_1(q)$ often leads to multiple, and very similar, instances of the same object, $\cN_2(q)$ provides higher variability of the negative examples, see Figure~\ref{fig:negatives}. While positives examples are fixed during the whole training process, hard negatives depend on the current CNN parameters and are re-mined multiple times per epoch. 

\setlength{\fboxsep}{0pt}%
\setlength{\fboxrule}{2pt}%

\section{Experiments}
\vspace{-5pt}
In this section we discuss implementation details of our training, evaluate different components of our method, and compare to the state of the art. 

\vspace{-5pt}
\subsection{Training setup and implementation details}
Our training samples are derived from the dataset used in the work of Schonberger~\etal~\cite{SRCF15}, which consists of 7.4 million images downloaded from Flickr using keywords of popular landmarks, cities and countries across the world.
The clustering procedure~\cite{CM10a} gives $19,546$ images to serve as query seeds. 
The extensive retrieval-SfM reconstruction~\cite{RSJFCM16} of the whole dataset results in $1,474$ reconstructed 3D models. 
Removing overlapping models leaves us with $713$ 3D models containing $163,671$ unique images from the initial dataset.
The initial dataset contained on purpose all images of Oxford5k and Paris6k datasets. 
In this way, we are able to exclude 98 clusters that contain any image (or their near duplicates) from these test datasets.

The largest model has $11,042$ images, while the smallest has $25$.
We randomly select $551$ models ($133,659$ images) for training and $162$ ($30,012$) for validation. 
The number of training queries per cluster is 10\% of the cluster size for clusters of 300 or less images, or 30 images for larger clusters. A total number of $5,974$ images is selected for training queries, and $1,691$ for validation queries.

Each training and validation tuple contains $1$ query, $1$ positive and $5$ negative images.
The pool of candidate positives consists of $k=100$ images with closest camera centers to the query.
In particular, for method $m_3$, the inliers overlap threshold is $t_i=0.2$, and the scale change threshold $t_s=1.5$.
Hard negatives are re-mined $3$ times per epoch, \ie roughly every $2,000$ training queries. 
Given the chosen queries and the chosen positives, we further add 20 images per cluster to serve as candidate negatives during re-mining.
This constitutes a training set of $22,156$ images and it corresponds to the case that all 3D models are included for training.

To perform the fine-tuning as described in Section~\ref{sec:network}, we initialize by the convolutional layers of AlexNet~\cite{KSH12} or VGG~\cite{SZ14}.
We use learning rate equal to $0.001$, which is divided by $5$ every $10$ epochs, momentum $0.9$, weight decay $0.0005$, parameter $\tau$ for contrastive loss $0.7$, and batch size of $5$ training tuples.
All training images are resized to a maximum $362 \times 362$ dimensionality, while keeping the original aspect ratio.
Training is done for at most $30$ epochs and the best network is selected based on performance, measured via mean Average Precision (mAP)~\cite{PCISZ07}, on validation tuples. 

\vspace{-5pt}
\subsection{Test datasets and evaluation protocol}
We evaluate our approach on Oxford buildings~\cite{PCISZ07}, Paris~\cite{PCISZ08} and Holidays\footnote{We use the up-right version of Holidays dataset (rotated).}~\cite{JDS08} datasets.
First two are closer to our training data, while the last differentiates by containing similar scenes and not only man made objects or buildings. 
These are also combined with 100k distractors from Oxford100k to allow for evaluation at larger scale. 
The performance is measured via mAP. 
We follow the standard evaluation protocol for Oxford and Paris and crop the query images with the provided bounding box. 
The cropped image is fed as input to the CNN.
However, to deliver a direct comparison with other methods, we also evaluate queries generated by keeping all activations that fall into this bounding box~\cite{BL15,AGTPS15} when the full query image is used as input to the network.
We refer to the cropped images approach as \cropI and the cropped activations~\cite{BL15,AGTPS15} as \cropA. 
The dimensionality of the images fed into the CNN is limited to $1024 \times 1024$ pixels.
In our experiments, no vector post-processing is applied if not otherwise stated.

\begin{figure}[t]
\centering
\input{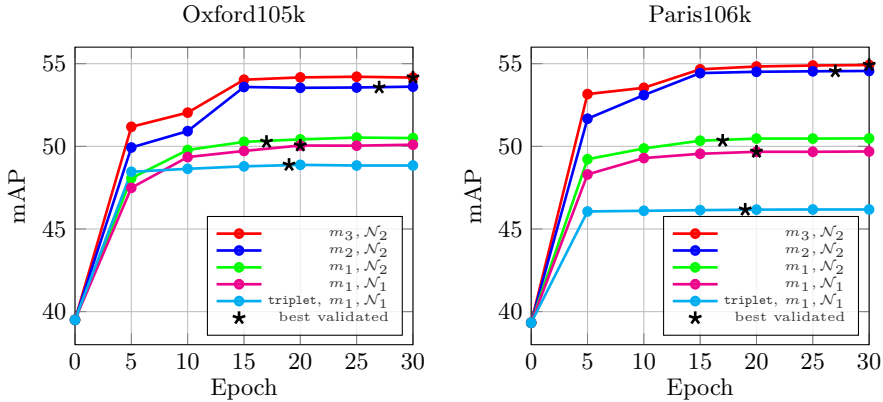}
\begin{tabular}{cc}
\extfig{posnegoxf}{ 
\begin{tikzpicture}
\begin{axis}[%
	width=0.5\linewidth,
	height=0.45\linewidth,
	xlabel={Epoch},
	ylabel={mAP},
	title={Oxford105k},
	legend pos=south east,
    legend style={cells={anchor=east}, font =\tiny, fill opacity=0.8, row sep=-2.5pt},	
    ymax = 56,
    ymin = 38,
    xmin = 0,
    xmax = 30,
    grid=both,
    xtick={0,5,...,30},    
    xticklabels={0,5,...,30},
    ytick={40, 45, ..., 55},
 	 y label style={at={(axis description cs:.1,.5)}},
 	 x label style={at={(axis description cs:.5,.05)}} 	
]

	\addplot[color=red,     solid, mark=*,  mark size=1.5, line width=1.0] table[x=epoch, y expr={100*\thisrow{siamrelaxed}}]       \posnegoxf;\leg{$m_3,\cN_2$};
	\addplot[color=blue,    solid, mark=*,  mark size=1.5, line width=1.0] table[x=epoch, y expr={100*\thisrow{siamninl}}]          \posnegoxf;\leg{$m_2,\cN_2$};
	\addplot[color=green,   solid, mark=*,  mark size=1.5, line width=1.0] table[x=epoch, y expr={100*\thisrow{siammac}}]           \posnegoxf;\leg{$m_1,\cN_2$};
	\addplot[color=magenta, solid, mark=*,  mark size=1.5, line width=1.0] table[x=epoch, y expr={100*\thisrow{siammacnegoff}}]     \posnegoxf;\leg{$m_1,\cN_1$};
	\addplot[color=cyan,    solid, mark=*,  mark size=1.5, line width=1.0] table[x=epoch, y expr={100*\thisrow{tripletmacnegoff}}]  \posnegoxf;\leg{\texttt{triplet}, $m_1,\cN_1$};
	\addplot[color=black, mark=star, only marks, mark size = 2.5, line width = 1] coordinates {(30,54.16)}; \leg{best validated};
	\addplot[color=black, mark=star, only marks, mark size = 2.5, line width = 1] coordinates {(27,53.56)};
	\addplot[color=black, mark=star, only marks, mark size = 2.5, line width = 1] coordinates {(17,50.27)};
	\addplot[color=black, mark=star, only marks, mark size = 2.5, line width = 1] coordinates {(20,50.05)};
	\addplot[color=black, mark=star, only marks, mark size = 2.5, line width = 1] coordinates {(19,48.88)};

\end{axis}
\end{tikzpicture}
}
&
\extfig{posnegpar}{ 
\begin{tikzpicture}
\begin{axis}[%
	width=0.5\linewidth,
	height=0.45\linewidth,
	xlabel={Epoch},
	ylabel={mAP},
	title={Paris106k},
	legend pos=south east,
    legend style={cells={anchor=east}, font =\tiny, fill opacity=0.8, row sep=-2.5pt},	
    ymax = 56,
    ymin = 38,
    xmin = 0,
    xmax = 30,
    grid=both,
    xtick={0,5,...,30},    
    xticklabels={0,5,...,30},
    ytick={40,45,...,55},
 	 y label style={at={(axis description cs:.1,.5)}},
 	 x label style={at={(axis description cs:.5,.05)}} 	    
]
	\addplot[color=red,     solid, mark=*,  mark size=1.5, line width=1.0] table[x=epoch, y expr={100*\thisrow{siamrelaxed}}]       \posnegpar;\leg{$m_3,\cN_2$};
	\addplot[color=blue,    solid, mark=*,  mark size=1.5, line width=1.0] table[x=epoch, y expr={100*\thisrow{siamninl}}]          \posnegpar;\leg{$m_2,\cN_2$};
	\addplot[color=green,   solid, mark=*,  mark size=1.5, line width=1.0] table[x=epoch, y expr={100*\thisrow{siammac}}]           \posnegpar;\leg{$m_1,\cN_2$};
	\addplot[color=magenta, solid, mark=*,  mark size=1.5, line width=1.0] table[x=epoch, y expr={100*\thisrow{siammacnegoff}}]     \posnegpar;\leg{$m_1,\cN_1$};
	\addplot[color=cyan,    solid, mark=*,  mark size=1.5, line width=1.0] table[x=epoch, y expr={100*\thisrow{tripletmacnegoff}}]  \posnegpar;\leg{\texttt{triplet}, $m_1,\cN_1$};
	\addplot[color=black, mark=star, only marks, mark size = 2.5, line width = 1] coordinates {(30,54.92)}; \leg{best validated};
	\addplot[color=black, mark=star, only marks, mark size = 2.5, line width = 1] coordinates {(27,54.54)};
	\addplot[color=black, mark=star, only marks, mark size = 2.5, line width = 1] coordinates {(17,50.34)};
	\addplot[color=black, mark=star, only marks, mark size = 2.5, line width = 1] coordinates {(20,49.67)};
	\addplot[color=black, mark=star, only marks, mark size = 2.5, line width = 1] coordinates {(19,46.17)};
	
\end{axis}
\end{tikzpicture}
}
\end{tabular}
\vspace{-15pt}
\caption{Performance comparison of methods for positive and negative example selection. Evaluation is performed on AlexNet MAC 
on Oxford105k and Paris106k datasets. The plot shows the evolution of mAP with the number of training epochs. Epoch 0 corresponds to the off-the-shelf network. All approaches use contrastive loss, except if otherwise stated. The network with the best performance on the validation set is marked with $\star$.
\label{fig:posnegmethod}}
\end{figure}
\begin{figure}[t]
\vspace{10pt}
\centering
\input{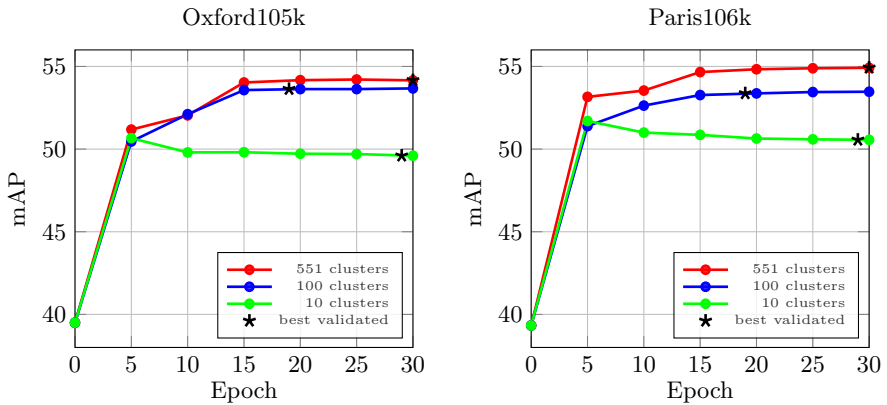}
\begin{tabular}{cc}
\extfig{clustnumoxf}{ 
\begin{tikzpicture}
\begin{axis}[%
	width=0.5\linewidth,
	height=0.45\linewidth,
	xlabel={Epoch},
	ylabel={mAP},
	title={Oxford105k},
	legend pos=south east,
    legend style={cells={anchor=east}, font =\tiny, fill opacity=0.8, row sep=-2.5pt},	
    ymax = 56,
    ymin = 38,
    xmin = 0,
    xmax = 30,
    grid=both,
    xtick={0,5,...,30},    
    xticklabels={0,5,...,30},
    ytick={40,45,...,55},
 	 y label style={at={(axis description cs:.1,.5)}},
 	 x label style={at={(axis description cs:.5,.05)}} 	    
]
	\addplot[color=red,     solid, mark=*,  mark size=1.5, line width=1.0] table[x=epoch, y expr={100*\thisrow{clustall}}]  \clustnumoxf;\leg{551 clusters};
	\addplot[color=blue,    solid, mark=*,  mark size=1.5, line width=1.0] table[x=epoch, y expr={100*\thisrow{clust100}}]  \clustnumoxf;\leg{100 clusters};
	\addplot[color=green,   solid, mark=*,  mark size=1.5, line width=1.0] table[x=epoch, y expr={100*\thisrow{clust10}}]   \clustnumoxf;\leg{10 clusters};
	\addplot[color=black, only marks, mark=star, mark size = 2.5, line width = 1] coordinates {(30,54.16)}; \leg{best validated};
	\addplot[color=black, only marks, mark=star, mark size = 2.5, line width = 1] coordinates {(19,53.63)};
	\addplot[color=black, only marks, mark=star, mark size = 2.5, line width = 1] coordinates {(29,49.60)};
	
\end{axis}
\end{tikzpicture}
}
&
\extfig{clustnumpar}{ 
\begin{tikzpicture}
\begin{axis}[%
	width=0.5\linewidth,
	height=0.45\linewidth,
	xlabel={Epoch},
	ylabel={mAP},
	title={Paris106k},
	legend pos=south east,
    legend style={cells={anchor=east}, font =\tiny, fill opacity=0.8, row sep=-2.5pt},	
    ymax = 56,
    ymin = 38,
    xmin = 0,
    xmax = 30,
    grid=both,
    xtick={0,5,...,30},    
    xticklabels={0,5,...,30},
    ytick={40,45,...,55},
 	 y label style={at={(axis description cs:.1,.5)}},
 	 x label style={at={(axis description cs:.5,.05)}} 	    
]
	\addplot[color=red,     solid, mark=*,  mark size=1.5, line width=1.0] table[x=epoch, y expr={100*\thisrow{clustall}}]  \clustnumpar;\leg{551 clusters};
	\addplot[color=blue,    solid, mark=*,  mark size=1.5, line width=1.0] table[x=epoch, y expr={100*\thisrow{clust100}}]  \clustnumpar;\leg{100 clusters};
	\addplot[color=green,   solid, mark=*,  mark size=1.5, line width=1.0] table[x=epoch, y expr={100*\thisrow{clust10}}]   \clustnumpar;\leg{10 clusters};
	\addplot[color=black, mark=star, only marks, mark size = 2.5, line width = 1] coordinates {(30,54.92)}; \leg{best validated};
	\addplot[color=black, mark=star, only marks, mark size = 2.5, line width = 1] coordinates {(19,53.37)};
	\addplot[color=black, mark=star, only marks, mark size = 2.5, line width = 1] coordinates {(29,50.56)};
	
\end{axis}
\end{tikzpicture}
}
\end{tabular}
\vspace{0pt}
\caption{Influence of the number of 3D models used for CNN fine-tuning. Performance is evaluated on AlexNet MAC on Oxford105k and Paris106k datasets using 10, 100 and 551 (all available) 3D models. The network with the best performance on the validation set is marked with $\star$.
\label{fig:clusternumber}
\vspace{10pt}}
\end{figure}
\subsection{Results on image retrieval}
\paragraph{Learning.}
We evaluate the off-the-shelf CNN and our fine-tuned ones after different number of training epochs. 
Our different methods for positive and negative selection are evaluated independently in order to decompose the benefit of each ingredient. 
Finally, we also perform a comparison with the triplet loss~\cite{AGTPS15}, trained on exactly the same training data as the ones used for our architecture with the contrastive loss. 
Results are presented in Figure~\ref{fig:posnegmethod}.
The results show that positive examples with larger view point variability, and negative examples with higher content variability, both acquire a consistent increase in the performance. 
The triplet loss\footnote{The margin parameter for the triplet loss is set equal to 0.1~\cite{AGTPS15}.} appears to be inferior in our context; we observe oscillation of the error in the validation set from early epochs, which implies over-fitting. 
In the rest of the paper, we adopt the $m_3,\cN_2$ approach.

\paragraph{Dataset variability.}
We perform fine-tuning by using a subset of the available 3D models. 
Results are presented in Figure~\ref{fig:clusternumber} with 10, 100 and 551 (all available) clusters, while keeping the amount of training data, \ie training queries, fixed.
In the case of 10 and 100 models we use the largest ones, \ie ones with the highest number of images.
It is better to train with all 3D models due to the higher variability in the training set. 
Remarkably, significant increase in performance is achieved even with 10 or 100 models. 
However, the network is able to over-fit in the case of few clusters.
All models are utilized in all other experiments.

\paragraph{Learned projections.}
The PCA-whitening~\cite{JC12} (\pcawhiten) is shown to be essential in some cases of CNN-based descriptors~\cite{BSCL14,BL15,TSJ16}.
On the other hand, it is shown that on some of the datasets, the performance after \pcawhiten substantially drops compared with the raw descriptors (max pooling on Oxford5k~\cite{BL15}). 
We perform comparison of this traditional way of whitening  and our learned whitening (\cpl2), described in Section~\ref{ref:projections}.
Table~\ref{tab:postproc} shows results without post-processing and with the two different methods of whitening.
Our experiments confirm, that \pcawhiten often reduces the performance. In contrast to that, the proposed
\cpl2 achieves the best performance in most cases and is never the worst performing method. Compared to no post-processing baseline, \cpl2 reduces the performance twice for AlexNet, but the drop is negligible compared to the drop observed for \pcawhiten. For the VGG, the proposed \cpl2 {\em always} outperforms the no post-processing baseline.

Our unsupervised learning directly optimizes MAC when extracted from full images, however, we further apply the fine-tuned networks to construct R-MAC representation~\cite{TSJ16} with regions of three different scales.
It consists of extracting MAC from multiple sub-windows and then aggregating them. 
Directly optimizing R-MAC during learning is possible and could offer extra improvements, but this is left for future work. Despite the fact that R-MAC offers improvements due to the regional representation, in our experiments it is not always better than MAC, since the latter is optimized during the end-to-end learning.
We apply \pcawhiten on R-MAC as in~\cite{TSJ16}, that is, we whiten each region vector first and then aggregate. Performance is significantly higher in this way. In the case of our \cpl2, we directly whiten the final vector after aggregation, which is also faster to compute.
\begin{table}[t!]
\caption{Performance comparison of CNN vector post-processing: no post-processing, PCA-whitening~\cite{JC12} (\pcawhiten) and our learned whitening (\cpl2). No dimensionality reduction is performed. Fine-tuned AlexNet produces a 256D vector and fine-tuned VGG a 512D vector. The best performance highlighted in \b{bold}, the worst in \ww{blue}. The proposed method consistently performs either the best (18 out of 24 cases) or on par with the best method. On the contrary, \pcawhiten~\cite{JC12} often hurts the performance significantly. Best viewed in color.
\label{tab:postproc}}
\vspace{-10pt}
\footnotesize
\begin{center}
\setlength{\tabcolsep}{0.6mm}
\begin{tabular}{|c|c|c|c|c|c|c|c|c|c|c|c|c|c|}
    \hline
    \multirow{2}{*}{Net} & \multirow{2}{*}{Post} & 
    \multicolumn{2}{c|}{Oxf5k} & \multicolumn{2}{c|}{Oxf105k} &
    \multicolumn{2}{c|}{Par6k}  & \multicolumn{2}{c|}{Par106k} & 
    \multicolumn{2}{c|}{Hol} & \multicolumn{2}{c|}{Hol101k} \\ 
    \cline{3-14}
    & & \tiny{MAC} & \tiny{R-MAC} & \tiny{MAC} & \tiny{R-MAC} & \tiny{MAC} & \tiny{R-MAC} & \tiny{MAC} 
    & \tiny{R-MAC} & \tiny{MAC} & \tiny{R-MAC} & \tiny{MAC} & \tiny{R-MAC}\\
    \hline\hline
    \multirow{3}{*}{Alex} 
    & --        	& 60.2 			& \ww{53.9}		& \b{54.2}      & \ww{46.4}		& 67.5 			& \ww{70.2}		& \b{54.9}      & \ww{58.4}		& 73.1 			& \ww{77.3}		& 61.6 			& \ww{67.1}	\\ 
	& \pcawhiten    & \ww{56.9}		& 60.0 			& \ww{44.1} 	& 48.4 			& \ww{64.3}		& \b{75.1}      & \ww{46.8}		& 61.7 			& \ww{73.0}		& \b{81.7}      & \ww{59.4}		& 70.4 		\\ 
	& \cpl2      	& \b{62.2}      & \b{62.5}      & 52.8 			& \b{53.2}      & \b{68.9}      & 74.4 		    & 54.7 			& \b{61.8}      & \b{76.2}      & 81.5 			& \b{63.8}      & \b{70.8}      \\ 
	\hline\hline
	\multirow{3}{*}{VGG} 
    & --        	& 78.7 			& \ww{70.1}		& 72.7 			& \ww{63.1}		& \ww{77.1}		& \ww{78.1}      & 69.6 			& \ww{70.4}		& \ww{76.9}		& \ww{80.0}		& 65.3 			& \ww{68.8} 	     \\ 
	& \pcawhiten    & \ww{76.1}		& 76.3 			& \ww{68.9}		& 68.5 			& 79.0 	        & \b{84.5}      & \ww{69.1}		& \b{77.1}      & 77.1 			& 82.3 			& \ww{63.6}		& 71.0 		 \\ 
	& \cpl2      	& \b{79.7}      & \b{77.0}      & \b{73.9}      & \b{69.2}	    & \b{82.4}      & 83.8          & \b{74.6}      & 76.4 	        & \b{79.5}      & \b{82.5}	    & \b{67.0}      & \b{71.5}      \\ 
	\hline
\end{tabular}
\end{center}
\end{table}

\begin{figure}[t!]
\centering
\input{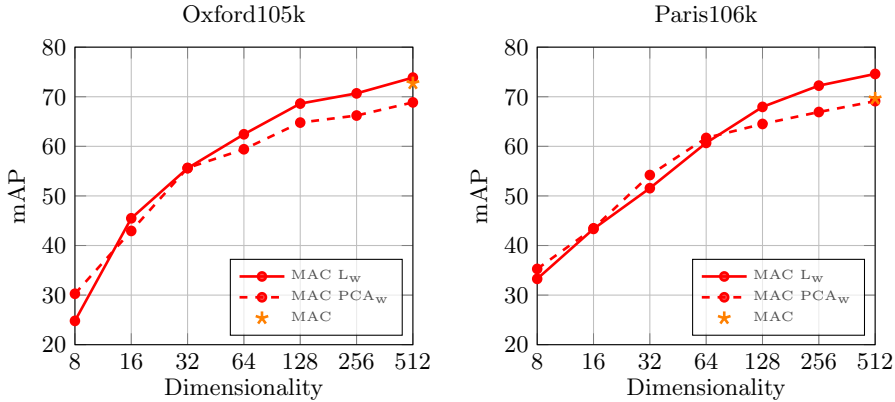}
\begin{tabular}{cc}
\extfig{dimRedOxf}{ 
\begin{tikzpicture}
\begin{semilogxaxis}[%
	width=0.5\linewidth,
	height=0.45\linewidth,
	xlabel={Dimensionality},
	ylabel={mAP},
	title={Oxford105k},
	legend pos=south east,
    legend style={cells={anchor=west}, font =\tiny, fill opacity=0.7, row sep=-1pt},	
    ymax = 80,
    ymin = 20,
    xmin = 8,
    xmax = 512,
    grid=both,
    xtick={8,16,32,64,128,256,512},    
    xticklabels={8,16,32,64,128,256,512},
    ytick={20,30,...,80},
 	 y label style={at={(axis description cs:.1,.5)}},
 	 x label style={at={(axis description cs:.5,.05)}} 	        
]
	\addplot[color=red,    solid,  mark=*,  mark size=1.5, line width=1.0] table[x=D, y expr={100*\thisrow{macVGGp}}]  \dimRedOxf;\leg{MAC \cpl2};
	\addplot[color=red,    dashed, mark options={solid}, mark=*,  mark size=1.5, line width=1.0] table[x=D, y expr={100*\thisrow{macVGGw}}]  \dimRedOxf;\leg{MAC \pcawhiten};
	\addplot[color=orange, mark=star, only marks, mark size = 2.5, line width = 1] coordinates {(512,72.71)}; \leg{MAC};
	
\end{semilogxaxis}
\end{tikzpicture}
}
&
\extfig{dimRedPar}{ 
\begin{tikzpicture}
\begin{semilogxaxis}[%
	width=0.5\linewidth,
	height=0.45\linewidth,
	xlabel={Dimensionality},
	ylabel={mAP},
	title={Paris106k},
	legend pos=south east,
    legend style={cells={anchor=west}, font =\tiny, fill opacity=0.7, row sep=-1pt},	
    ymax = 80,
    ymin = 20,
    xmin = 8,
    xmax = 512,
    grid=both,
    xtick={8,16,32,64,128,256,512},    
    xticklabels={8,16,32,64,128,256,512},
    ytick={20,30,...,80},
 	 y label style={at={(axis description cs:.1,.5)}},
 	 x label style={at={(axis description cs:.5,.05)}} 	        
]
	\addplot[color=red,    solid,  mark=*,  mark size=1.5, line width=1.0] table[x=D, y expr={100*\thisrow{macVGGp}}]  \dimRedPar;\leg{MAC \cpl2};
	\addplot[color=red,    dashed, mark options={solid}, mark=*,  mark size=1.5, line width=1.0] table[x=D, y expr={100*\thisrow{macVGGw}}]  \dimRedPar;\leg{MAC \pcawhiten};
	\addplot[color=orange, mark=star, only marks, mark size = 2.5, line width = 1] coordinates {(512,69.63)}; \leg{MAC};
	
\end{semilogxaxis}
\end{tikzpicture}
}
\end{tabular}
\vspace{-15pt}
\caption{Performance comparison of the dimensionality reduction performed by \pcawhiten and our \cpl2 on fine-tuned VGG MAC on Oxford105k and Paris106k datasets.
\label{fig:dimreduce}}
\end{figure}

\vspace{10pt}
\paragraph{Dimensionality reduction.}
We compare dimensionality reduction performed with \pcawhiten~\cite{JC12} and with our \cpl2. The performance for varying descriptor dimensionality is plotted in Figure~\ref{fig:dimreduce}. The plots suggest that \cpl2 works better in higher dimensionalities, while \pcawhiten works slightly better for the lower ones. Remarkably, MAC reduced down to 16D outperforms state-of-the-art on BoW-based 128D compact codes~\cite{RJC15} on Oxford105k ($45.5$ vs $41.4$). Further results on very short codes can be found in Table~\ref{tab:stateofart}. 

\vspace{10pt}
\paragraph{Over-fitting and generalization.}
In all experiments, all clusters including any image (not only query landmarks) from Oxford5k or Paris6k datasets are removed. 
To evaluate whether the network tends to over-fit to the training data or to generalize, we repeat the training, this time using all 3D reconstructions, including those of Oxford and Paris landmarks.
The same amount of training queries is used for a fair comparison.
We observe negligible difference in the performance of the network on Oxford and Paris evaluation results, \ie the difference in mAP was on average $+0.3$ over all testing datasets. 
We conclude that the network generalizes well and is relatively insensitive to over-fitting. 

\newcommand{\ebA}{\texttt{\bf{A}}}
\newcommand{\ebV}{\texttt{\bf{V}}}
\newcommand{\ebf}{\texttt{\bf{f}}}
\newcommand{\our}{$\boldsymbol{\star}$\xspace}

\begin{table}[t!]
\caption{Performance comparison with the state of the art. Results reported with the use of AlexNet or VGG are marked by (\ebA) or (\ebV), respectively. Use of fine-tuned network is marked by (\ebf), otherwise the off-the-shelf network is implied. \mbox{D:~Dimensionality}. Our methods are marked with \our and they are always accompanied by \cpl2. New state of the art highlighted in \nb{red}, surpassed state of the art in \ob{bold}, state of the art that retained the title in \protect\sb{outline}, and our methods that outperform previous state of the art on a \fcolorbox{gray!15}{gray!15}{gray background}. Best viewed in color.
\label{tab:stateofart}}
\vspace{-5pt}
\footnotesize
\begin{center}
\setlength{\tabcolsep}{0.4mm}
\begin{tabular}{|l@{}r|c|c|c|c|c|c|c|c|c|c|c|}
    \hline
     \multicolumn{2}{|c|}{\multirow{2}{*}{Method}}
     & \multirow{2}{*}{D} & 
     \multicolumn{2}{c|}{Oxf5k} & \multicolumn{2}{c|}{Oxf105k} &
     \multicolumn{2}{c|}{Par6k}  & \multicolumn{2}{c|}{Par106k} & 
     Hol & Hol \\ 
     \cline{4-11}
    & & & \tiny{\cropI} & \tiny{\cropA} & \tiny{\cropI} & \tiny{\cropA} & \tiny{\cropI} & \tiny{\cropA} & \tiny{\cropI} & \tiny{\cropA} &  & \scriptsize 101k \\ 
    \hline 
    \multicolumn{13}{c}{~} \\[-2ex]
    \hline
    \multicolumn{13}{|c|}{Compact representations} \\ \hline 
    mVoc/BoW~\cite{RJC15}&                  & 128 & 48.8         & --           & 41.4         & --           & --           & --           & --           & --           & 65.6         & --        \\
    Neural codes$^\dagger$~\cite{BSCL14} 
                           &(\ebf\ebA)      & 128 & --           & \ob{55.7}    & --           & \ob{52.3}    & --           & --           & --           & --           & \ob{78.9}    & --        \\
    MAC$^\ddagger$         &(\ebV)          & 128 & 53.5         & \ob{55.7}    & 43.8         & 45.6         & 69.5         & \ob{70.6}    & 53.4         & \ob{55.4}    & 72.6         & \ob{56.7} \\
    CroW~\cite{KMO15}      &(\ebV)          & 128 & \ob{59.2}    & --           & \ob{51.6}    & --           & \ob{74.6}    & --           & \ob{63.2}    & --           & --           & --        \\ 
    \our MAC               &(\ebf\ebV)      & 128 & \bo\nb{75.8} & \bo\nb{76.8} & \bo\nb{68.6} & \bo\nb{70.8} & \bo77.6      & \bo78.8      & \bo68.0      & \bo69.0      & 73.2         & \bo58.8      \\
    \our R-MAC             &(\ebf\ebV)      & 128 & \bo72.5      & \bo76.7      & \bo64.3      & \bo69.7      & \bo\nb{78.5} & \bo\nb{80.3} & \bo\nb{69.3} & \bo\nb{71.2} & \bo\nb{79.3} & \bo\nb{65.2} \\
    \hline
    MAC$^\ddagger$         &(\ebV)          & 256 & 54.7         & 56.9         & 45.6         & 47.8         & 71.5         & 72.4    & 55.7         & \ob{57.3}    & 76.5         & \ob{61.3} \\
    SPoC~\cite{BL15}       &(\ebV)          & 256 & --           & 53.1         & --           & \ob{50.1}    & --           & --           & --           & --           & 80.2         & --        \\
    R-MAC~\cite{TSJ16}     &(\ebA)          & 256 & 56.1         & --           & 47.0         & --           & 72.9         & --           & 60.1         & --           & --           & --        \\    
    CroW~\cite{KMO15}      &(\ebV)          & 256 & \ob{65.4}    & --           & \ob{59.3}    & --           & \ob{77.9}    & --           & \ob{67.8}    & --           & 83.1         & --        \\ 
    NetVlad~\cite{AGTPS15} &(\ebV)          & 256 & --           & 55.5         & --           & --           & --           & 67.7         & --           & --           & \sb{86.0}    & --        \\ 
    NetVlad~\cite{AGTPS15} &(\ebf\ebV)      & 256 & --           & \ob{63.5}    & --           & --           & --           & \ob{73.5}         & --           & --           & 84.3         & --        \\
    \our MAC               &(\ebf\ebA)      & 256 & 62.2         & \bo65.4      & 52.8         & \bo58.0      & 68.9         & 72.2         & 54.7         & \bo58.5      & 76.2         & \bo63.8      \\
    \our R-MAC             &(\ebf\ebA)      & 256 & 62.5         & \bo68.9      & 53.2         & \bo61.2      & 74.4         & \bo76.6      & 61.8         & \bo64.8      & 81.5         & \bo\nb{70.8} \\
    \our MAC               &(\ebf\ebV)      & 256 & \bo\nb{77.4} & \bo\nb{78.2} & \bo\nb{70.7} & \bo\nb{72.6} & \bo80.8      & \bo81.9      & \bo72.2      & \bo73.4      & 77.3         & \bo62.9      \\
    \our R-MAC             &(\ebf\ebV)      & 256 & \bo74.9      & \bo\nb{78.2} & \bo67.5      & \bo72.1      & \bo\nb{82.3} & \bo\nb{83.5} & \bo\nb{74.1} & \bo\nb{75.6} & 81.4         & \bo69.4      \\
    \hline
    MAC$^\ddagger$         &(\ebV)          & 512 & 56.4         & \ob{58.3}    & 47.8         & \ob{49.2}    & 72.3         & \ob{72.6}    & 58.0         & \ob{59.1}    & 76.7         & \ob{62.7} \\
    R-MAC~\cite{TSJ16}     &(\ebV)          & 512 & 66.9         & --           & 61.6         & --           & \ob{83.0}    & --           & \ob{75.7}    & --           & --           & --        \\     
    CroW~\cite{KMO15}      &(\ebV)          & 512 & \ob{68.2}    & --           & \ob{63.2}    & --           & 79.6         & --           & 71.0         & --           & 84.9         & --        \\    
    \our MAC               &(\ebf\ebV)      & 512 & \bo\nb{79.7} & \bo80.0      & \bo\nb{73.9} & \bo\nb{75.1} & 82.4         & \bo82.9      & 74.6         & \bo75.3      & 79.5         & \bo67.0      \\
    \our R-MAC             &(\ebf\ebV)      & 512 & \bo77.0      & \bo\nb{80.1} & \bo69.2      & \bo74.1      & \bo\nb{83.8} & \bo\nb{85.0} & \bo\nb{76.4} & \bo\nb{77.9} & 82.5         & \bo\nb{71.5} \\
    \hline 
    \multicolumn{13}{c}{~} \\[-2ex]
    \hline
    \multicolumn{13}{|c|}{Extreme short codes} \\ \hline 
    Neural codes$^\dagger$~\cite{BSCL14} 
                           &(\ebf\ebA)      & 16  & --           & \ob{41.8}    & --           & \ob{35.4}    & --           & --           & --           & --           & \sb{60.9}    & --        \\
    \our MAC               &(\ebf\ebV)      & 16  & \bo\nb{56.2} & \bo\nb{57.4} & \bo\nb{45.5} & \bo\nb{47.6} & \bo57.3      & \bo62.9      & \bo43.4      & \bo48.5      & 51.3         & \bo25.6      \\
    \our R-MAC             &(\ebf\ebV)      & 16  & \bo46.9      & \bo52.1      & \bo37.9      & \bo41.6      & \bo\nb{58.8} & \bo\nb{63.2} & \bo\nb{45.6} & \bo\nb{49.6} & 54.4         & \bo\nb{31.7} \\
    \hline    
    Neural codes$^\dagger$~\cite{BSCL14} 
                           &(\ebf\ebA)      & 32  & --           & \ob{51.5}    & --           & \ob{46.7}    & --           & --           & --           & --           & \sb{72.9}    & --        \\
    \our MAC               &(\ebf\ebV)      & 32  & \bo\nb{65.3} & \bo\nb{69.2} & \bo\nb{55.6} & \bo\nb{59.5} & \bo\nb{63.9} & \bo\nb{69.5} & \bo51.6      & \bo\nb{56.3} & 62.4         & \bo41.8      \\
    \our R-MAC             &(\ebf\ebV)      & 32  & \bo58.4      & \bo64.2      & \bo50.1      & \bo55.1      & \bo\nb{63.9} & \bo67.4      & \bo\nb{52.7} & \bo55.8      & 68.0         & \bo\nb{49.6} \\
    \hline 
    \multicolumn{13}{c}{~} \\[-2ex]
    \hline
    \multicolumn{13}{|c|}{Re-ranking (R) and query expansion (QE)} \\ \hline 
    BoW(1M)+QE~\cite{CMPM11}      &             & --  & 82.7         & --           & 76.7         & --           & 80.5         & --           & 71.0         & --           & --           & --   \\
    BoW(16M)+QE~\cite{MPCM13}     &        & -- & 84.9         & --           & 79.5         & --           & 82.4         & --           & 77.3         & --           & --           & --   \\
    HQE(65k)~\cite{TJ14}           &             & -- & \sb{88.0}    & --           & \sb{84.0}    & --           & 82.8         & --           & --           & --           & --           & --   \\
    R-MAC+R+QE~\cite{TSJ16}  &(\ebV)       & 512 & 77.3         & --           & 73.2         & --           & \sb{86.5}    & --           & \sb{79.8}    & --           & --           & --   \\ 
    CroW+QE~\cite{KMO15}      &(\ebV)       & 512 & 72.2         & --           & 67.8         & --           & 85.5         & --           & 79.7         & --           & --           & --   \\       
    \our MAC+R+QE            &(\ebf\ebV)   & 512 & 85.0         & \bo\nb{85.4} & 81.8         & \bo\nb{82.3} & \bo\nb{86.5} & \bo\nb{87.0} & 78.8         & \bo79.6      & --           & --   \\
    \our R-MAC+R+QE          &(\ebf\ebV)   & 512 & 82.9         & \bo84.5      & 77.9         & \bo80.4      & 85.6         & \bo86.4      & 78.3         & \bo\nb{79.7} & --           & --   \\
    \hline
\end{tabular}
\end{center}
\vspace{-5pt}
\scriptsize
$^\dagger$:~Full images are used as queries making the results not directly comparable on Oxford and Paris. \\
$^\ddagger$:~Our evaluation of MAC with \pcawhiten as in~\cite{TSJ16} with the off-the-shelf network.
\end{table}

\paragraph{Comparison with the state of the art.}
We extensively compare our results with the state-of-the-art performance on compact image representations and extremely short codes. The results for MAC and R-MAC with the fine-tuned networks are summarized together with previously published results in Table~\ref{tab:stateofart}. The proposed methods outperform the state of the art on Paris and Oxford datasets, with and without distractors with all 16D, 32D, 128D, 256D, and 512D descriptors. On Holidays dataset, the Neural codes~\cite{BSCL14} win the extreme short code category, while off-the-shelf NetVlad performs the best on 256D and higher. 

We additionally combine MAC and R-MAC with recent localization method for re-ranking~\cite{TSJ16} to further boost the performance. Our scores compete with state-of-the-art systems based on local features and query expansion. These have much higher memory needs and larger query times. 

Observations on the recently published NetVLAD~\cite{AGTPS15}:
(1) After fine-tuning, NetVLAD performance drops on Holidays, while our training improves off-the-shelf results on all datasets.
(2) Our 32D MAC descriptor has comparable performance to 256D NetVLAD  on Oxford5k (ours $69.2$ vs NetVLAD $63.5$), and on Paris6k (ours $69.5$ vs NetVLAD $73.5$).

\vspace{-2pt}
\section{Conclusions}
%
We addressed fine-tuning of CNN for image retrieval. The training data are selected from an automated 3D reconstruction system applied on a large unordered photo collection. The proposed method does not require any manual annotation and yet outperforms the state of the art on a number of standard benchmarks for wide range (16 to 512) of descriptor dimensionality. The achieved results are reaching the level of the best systems based on local features with spatial matching and query expansion, while being faster and requiring less memory. Training data, fine-tuned networks and evaluation code are publicly available\footnote{\href{http://cmp.felk.cvut.cz/~radenfil/projects/siamac.html}{http://cmp.felk.cvut.cz/\~{}radenf{}i{}l/projects/siamac.html}}.

\clearpage

\paragraph{Acknowledgment}. Work was supported by the MSMT LL1303 ERC-CZ grant.

\bibliographystyle{splncs}
\bibliography{egbib}
\end{document}